\documentclass[
twocolumn,
hf,
]{ceurart}

\usepackage{listings}
\usepackage{amsmath}
\usepackage{caption}
\usepackage{url}
\begin{document}

\copyrightyear{2025}
\copyrightclause{Copyright for this paper by its authors.
  Use permitted under Creative Commons License Attribution 4.0
  International (CC BY 4.0).}

% \conference{Woodstock'22: Symposium on the irreproducible science,
%   June 07--11, 2022, Woodstock, NY}

\conference{6th Workshop on Patent Text Mining and Semantic Technologies (PatentSemTech) 2025}

\title{Hierarchical Multi-Positive Contrastive Learning for Patent Image Retrieval}

% \tnotetext[1]{You can use this document as the template for preparing your
%   publication. We recommend using the latest version of the ceurart style.}


\author[1]{Kshitij Kavimandan}[%
orcid=0009-0002-3760-5882,
email=kshitijk3188@gmail.com,
]
\cormark[1]

\author[2]{Angelos Nalmpantis}[%
orcid=0000-0002-1505-4656,
email=a.nalmpantis@tkh.ai,
]

\author[1]{Emma Beauxis-Aussalet}[%
orcid=0000-0002-4657-892X,
email=e.m.a.l.beauxisaussalet@vu.nl,
]

\author[2]{Robert-Jan Sips}[%
orcid=0000-0002-2316-7183,
email=r.sips@tkh.ai,
]

\address[1]{Vrije Universiteit Amsterdam, Amsterdam, The Netherlands}
\address[2]{TKH AI, Amsterdam, The Netherlands}

% %% Footnotes
\cortext[1]{Corresponding author.}
% \fntext[1]{These authors contributed equally.}

\begin{abstract}
Patent images are technical drawings that convey information about a patent's innovation. Patent image retrieval systems aim to search in vast collections and retrieve the most relevant images. Despite recent advances in information retrieval, patent images still pose significant challenges due to their technical intricacies and complex semantic information, requiring efficient fine-tuning for domain adaptation. Current methods neglect patents' hierarchical relationships, such as those defined by the Locarno International Classification (LIC) system, which groups broad categories (e.g., ``furnishing'') into subclasses (e.g., ``seats'' and ``beds'') and further into specific patent designs. In this work, we introduce a hierarchical multi-positive contrastive loss that leverages the LIC's taxonomy to induce such relations in the retrieval process. Our approach assigns multiple positive pairs to each patent image within a batch, with varying similarity scores based on the hierarchical taxonomy. Our experimental analysis with various vision and multimodal models on the DeepPatent2 dataset shows that the proposed method enhances the retrieval results. Notably, our method is effective with low-parameter models, which require fewer computational resources and can be deployed on environments with limited hardware.
\end{abstract}


\maketitle

\section{Introduction}
Patent images are technical drawings that illustrate the novelty of a patent, often conveying their details more effectively than natural language written in text \cite{kucer2022deeppatent}. Thereby, technical patent reports are typically accompanied by multiple images capturing different aspects of the invention. With the rapidly growing volume of patents, efficient patent image retrieval systems are becoming an essential component for searching these vast collections. 

Many advances in information retrieval have been largely driven by the power of attention based models \cite{vaswani2017attention, devlin2019bert} and the knowledge acquired during extensive pretraining phases, mainly focused on the language domain. While similar models, such as Vision Transformer (ViT) \cite{dosovitskiy2021an} and ResNet \cite{he2016deep}, have provided remarkable results on a plethora of vision tasks, they still fall short when processing technical drawings since their pretraining mainly involves natural images \cite{geirhos2018imagenet}. In response, to address this domain shift, researchers have released specialized sketch datasets \cite{wang2019learning, sangkloy2016sketchy} that facilitate model fine-tuning on such images. Similarly, large scale datasets containing patent images have emerged to address their unique intricacies and enable the development of efficient patent image retrieval methods. 

DeepPatent \cite{kucer2022deeppatent} was the first large scale dataset designed for training and evaluating patent image retrieval systems, comprising over $350,000$ images across $45,000$ patents, enabling the development of PatentNet, which exhibited significant improvements in patent image retrieval. Additionally, several studies investigated the generation of synthetic text descriptions by leveraging the zero-shot capabilities of (vision) language models \cite{aubakirova2023patfig, lo2024largelanguagemodelinformed}, allowing the application of multimodal models, such as CLIP \cite{radford2021learning}, on patent image retrieval. Inspired by DeepPatent, DeepPatent2 \cite{ajayi2023deeppatent2} provided an extension of the dataset, scaling to more than $2.7$ million images with patents spanning from $2007$ to $2020$ while also incorporating additional metadata like the object's name. Despite the advances in patent image retrieval \cite{wang2023learning, higuchi2023patent}, many methodologies determine the relevance of images based on their association with the same patent. This criterion neglects the rich hierarchical taxonomies of patents that are defined by standardized classification systems. Such hierarchical similarities could potentially enhance the effectiveness of patent image retrieval systems.

\begin{figure*}
    \centering
    \includegraphics[width=\linewidth]{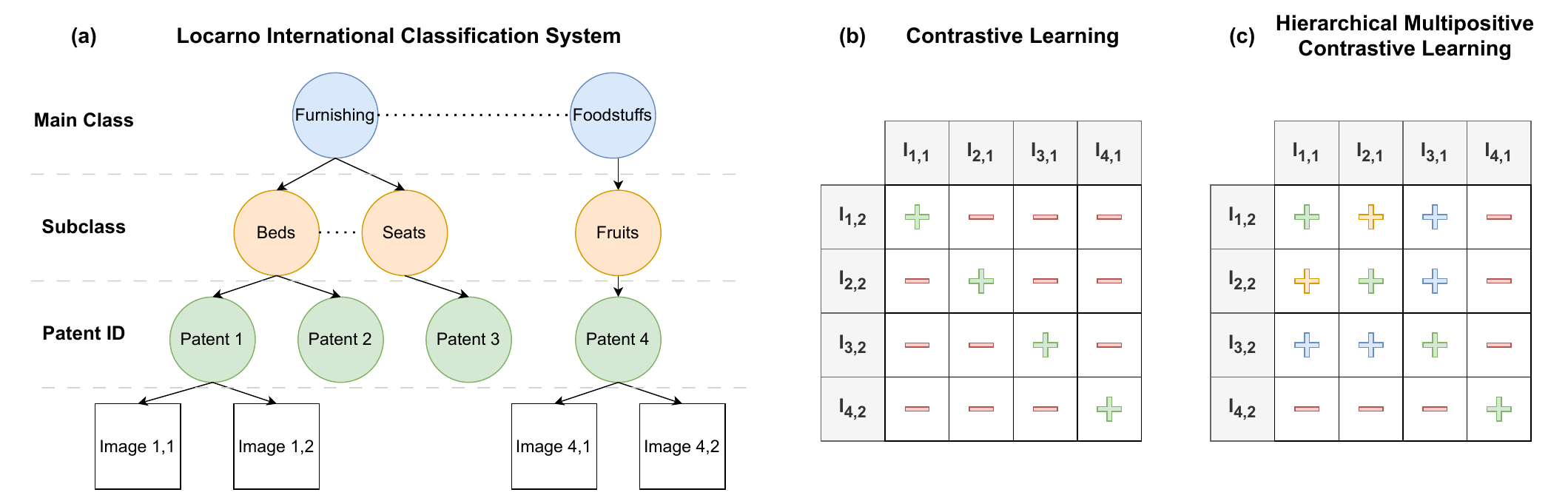}
   \caption{Overview of the Hierarchical Multi-Positive Contrastive Learning approach using the Locarno International Classification system. Figure (a) is an example of the hierarchical taxonomy of patents provided by the Locarno International Classification system. The images from Patent 2 and Patent 3 are omitted. Figure (b) presents how positive pairs within a batch are defined in conventional contrastive learning approaches. Figure (c) summarizes the proposed Hierarchical Multi-Positive Contrastive Learning approach, where the anchor images have multiple positive pairs within the batch, each assigned a different similarity score based on the hierarchical relationship from (a). $\text{I}_{i,j}$ denotes the $j$-th image that belongs to patent $i$.}
    \label{fig:lic_ct_hmct}
\end{figure*}

In this paper, we aim to address this limitation by leveraging the hierarchical taxonomy of patents as defined by the Locarno International Classification (LIC) system \cite{locarno}, which organizes industrial designs into a structured taxonomy consisting of $32$ main classes, each further divided into various subclasses. Figure~\ref{fig:lic_ct_hmct} provides an example of how patents are organized within this hierarchical taxonomy. For brevity, we omit illustrating all classes entailed in the LIC taxonomy. While many studies aim to capture the inherent hierarchical information of data, ranging from representation learning methods \cite{mettes2024hyperbolic, nalmpantis2023hierarchical} to specialized architectures \cite{liu2021swin}, it remains unclear how to properly leverage patents' hierarchical relations for improving patent image retrieval. 

To this end, we propose a hierarchical multi-positive contrastive learning method that explicitly integrates these hierarchical relations of patents into the training process. Our method extends upon previous works on patent image retrieval \cite{kucer2022deeppatent} and contrastive learning approaches \cite{radford2021learning, tian2023stablerep} by treating patent images of the same hierarchical main class, subclass and patent ID as positive with varying degrees of similarity. Figure~\ref{fig:lic_ct_hmct} compares conventional contrastive learning methods with the proposed approach. With the conventional method shown in Figure~\ref{fig:lic_ct_hmct}(b), each image is associated only with one positive pair that belongs to the same patent ID. In contrast, the proposed approach in Figure~\ref{fig:lic_ct_hmct}(c) respects the hierarchical taxonomy, assigning higher positive scores to images with finer taxonomic relationships. For example, two images from the same patent receive the highest positive score, reflecting their direct relationship. Images that belong only to the same Locarno subclass are assigned a slightly lower positive score, while those that share only the same Locarno main class receive an even lower score.

In our experimental analysis with various architectures, we demonstrate that our approach enhances the retrieval performance. Notably, the proposed method shows great effectiveness with low parameter models which can be deployed in resource constrained environments where computational efficiency is also crucial.

The rest of the paper is structured as follows. First, in Section~\ref{sec:methodology}, we formulate the proposed hierarchical multi-positive contrastive learning method for patents. Then, in Section~\ref{sec:experimental_setup}, we provide the details of the experimental setup, facilitating the reproducibility of our results. In Section~\ref{sec:experimental_results}, we report our findings and demonstrate the effectiveness of our approach. Finally, in Section~\ref{sec:conclusion}, we draw the conclusions of this study and discuss future directions.

\section{Methodology}
\label{sec:methodology}
To induce hierarchical relations among patents, we propose a Hierarchical Multi-Positive Contrastive Learning approach that leverages the hierarchical taxonomy provided by the LIC system. Our approach enables the model to align patent images of the same main class, subclass and patent ID incrementally closer in the embedding space. 

Let $X$ be a collection of patent images, $x \in X$ a sample image from the dataset and $z \in \mathbb{R}^d$ the corresponding image embedding provided by an image encoder. Considering a batch of $2K$ images that form $K$ positive pairs $(x_i, \Tilde{x}_i)$ and the embeddings of the anchor image $z_i$ and its positive pair $\Tilde{z}_i$, the contrastive loss \cite{radford2021learning, oord2019representationlearningcontrastivepredictive} is defined as:

\begin{equation}
\label{eq:ct}
    L_{i} = -\log\frac{\exp( \text{sim}(z_i, \Tilde{z}_i))}{\sum_{j=1}^{K}\exp(\text{sim}(z_i, \Tilde{z}_j))}
\end{equation}
where $\text{sim}(z_i, \Tilde{z}_j)$ indicates the cosine similarity between the two vector embeddings $z_i$ and $\Tilde{z}_j$.

The loss defined in Equation~\ref{eq:ct}, as well as similar losses employed in prior work, such as in PatentNet, are unable to properly capture the hierarchical relations of patents within the batch. In contrast, $L_i$ should accommodate multiple positive pairs for the anchor image $x_i$ and assign a different relevance score to each pair depending on their hierarchical relations within the LIC taxonomy.

Let $h_{ij}$ define the relevance score between two images $x_i$ and $\Tilde{x}_j$:

\begin{equation}
    h_{ij} = 
    \begin{cases}
        s_{p} & \text{if $x_i$ and $\Tilde{x}_j$ belong to the same patent ID} \\
        s_{s} & \text{if $x_i$ and $\Tilde{x}_j$ belong to the same subclass} \\
        s_{m} & \text{if $x_i$ and $\Tilde{x}_j$ belong to the same main class} \\
        0 & \text{otherwise}
    \end{cases}
\end{equation}
where $s_{p} > s_{s} > s_{m}$ are positive scalar values that reflect the importance of matching at different hierarchical levels. The function $h_{ij}$ assigns the highest relevance score to the most specific case (same patent ID) with progressively lower scores for broader relationships. Additionally, let $H_i$ be the normalization factor for the patent image $x_i$:

\begin{equation}
    H_i = \sum_{j=1}^{K} h_{ij}
\end{equation}
Then, the hierarchical multi-positive contrastive loss is defined as:
\begin{equation}
\label{eq:hct}
    L_{i} = - \sum_{j=1}^{K} \frac{h_{ij}}{H_i}\log\frac{\exp(\text{sim}(z_i, \Tilde{z}_j))}{\sum_{l=1}^{K}\exp(\text{sim}(z_i, \Tilde{z}_l))}
\end{equation}
This formulation enables the model to learn representations that align each image $x_i$ with multiple other images from the batch based on their hierarchical proximity within the LIC taxonomy.

In the case where the text description $t_i$ of the patent image $x_i$ is available, we can incorporate language supervision by adding an additional term to $L_i$:
\begin{equation}
\label{eq:language_loss}
    - \lambda \sum_{j=1}^{K} \frac{h_{ij}}{H_i}\log\frac{\exp(\text{sim}(z_i, y_j))}{\sum_{l=1}^{K}\exp(\text{sim}(z_i, y_{l}))}
\end{equation}
where $y_j$ denotes the embedding of the text description $t_j$ provided by a language encoder. The hyperparameter $\lambda$ is a weighting factor controlling the language supervision. 

Note that Equation~\ref{eq:ct} is a special case of Equation~\ref{eq:hct}. The two equations are equivalent when only a single positive pair exists with a score of 1, and all other pairs are assigned a score of 0.

While our implementation leverages the LIC system, this approach generalizes to other hierarchical classification systems, such as the Cooperative Patent Classification system. Alternative taxonomies can be seamlessly integrated by appropriately defining the scoring function $h_{ij}$ to reflect their specific hierarchical structures.

\section{Experimental Setup}
\label{sec:experimental_setup}

For conducting the experiments, we use the DeepPatent2 dataset \cite{ajayi2023deeppatent2} for the year 2007, which contains multiple images per patent along with the patent's code from the LIC system and a short textual description of the depicted object. The experimental setup is similar to \citet{kucer2022deeppatent}. We split the data using a $72.25/12.75/15$ ratio for training, validation and testing, respectively. In training, we sample $64$ patents, where for each we randomly pick $2$ images forming a positive pair based on the patent ID. For testing, we sample $2$ images from each patent, with each image being used individually to form a query. The rest of the patent images from the test set form the database used for searching. All images are reshaped with a resolution of $224\times224$. During training, we use the following augmentation techniques to avoid overfitting: horizontal flip with an applying probability of $0.3$, rotation by a maximum of $10$ degrees with a probability of $0.5$, and Gaussian noise with a probability of $0.2$. For testing, no augmentation methods are deployed. 

We conduct experiments with the ViT \cite{dosovitskiy2021an}, CLIP \cite{radford2021learning}, and ResNet \cite{he2016deep} architecture of different sizes. The vision models, ViT and ResNet, are initialized from a pretrained version on ImageNet, while CLIP models are pretrained using the dataset from \cite{radford2021learning}.

We use the AdamW optimizer \cite{loshchilov2018decoupled} with a learning rate of $0.0001$ and weight decay of $0.01$. All models are trained for 20 epochs until convergence with early stopping based on the validation set. Each experiment is repeated for multiple random seeds. For the ViT and ResNet models, we repeat the experiments for 5 different seeds, while for the CLIP models, which require more computational resources, we use $3$ different seeds. The temperature $\tau$ and the hyperparameter $\lambda$ are set to $0.1$ and $0.2$, respectively. For the scoring function $h_{ij}$, we set $s_{p}=1$, $s_{s}=0.35$ and $s_{m}=0.2$, emphasizing the patent ID level while still incorporating information from higher levels. These values offer a balanced performance across all levels and a fair comparison with the baselines that mainly focus on the patent ID level. Note that a different scoring function could be used depending on the significance of each hierarchical level for the use case at hand.

The models are evaluated using the mean Average Precision (mAP), the normalized Discounted Cumulative Gain (nDCG), the Top-K Mean Reciprocal Rank (MRR@K) and the Top-K Accuracy (Acc@K).

The experiments are conducted using PyTorch \cite{paszke2019pytorchimperativestylehighperformance}, PyTorch Lightning \cite{Falcon_PyTorch_Lightning_2019}, and the transformers library from Hugging Face \cite{wolf-etal-2020-transformers}. The training process of a model takes approximately $2.5$ hours on a single NVIDIA A100 GPU.

\section{Experimental Results}
\label{sec:experimental_results}

\begin{table}[t]
\centering
\footnotesize
\caption{Performance of the ResNet and ViT models on patent image retrieval. All results are averaged across $5$ runs with different seeds.}
\label{tab:resnet_vit_results}
\scalebox{0.78}{
\begin{tabular}{>{\columncolor{gray!10}}l|l|cc|cc|cc}
\toprule
\rowcolor{gray!20}
\textbf{Model} & \textbf{Method} & \multicolumn{2}{c|}{\textbf{Patent ID}} & \multicolumn{2}{c|}{\textbf{Subclass}} & \multicolumn{2}{c}{\textbf{Main Class}} \\
& & \textbf{mAP} & \textbf{nDCG} & \textbf{mAP} & \textbf{nDCG} & \textbf{mAP} & \textbf{nDCG} \\
\midrule
\rowcolor{gray!5}
\multicolumn{8}{l}{\textbf{ResNet Models}} \\
\midrule
\multirow{3}{*}{ResNet-18} 
    & Baseline     & 0.153 & 0.366 & 0.053 & 0.484 & 0.076 & 0.610 \\
ResNet-18 & CL       & 0.208 & 0.437 & 0.079 & 0.514 & 0.095 & 0.626 \\
    & HMCL   & \textbf{0.221} & \textbf{0.453} & \textbf{0.085} & \textbf{0.521} & \textbf{0.101} & \textbf{0.631} \\
\midrule
\multirow{3}{*}{ResNet-34} 
    & Baseline     & 0.152 & 0.366 & 0.058 & 0.494 & 0.080 & 0.613 \\
ResNet-34 & CL       & 0.212 & 0.447 & 0.086 & 0.523 & 0.100 & 0.629 \\
    & HMCL   & \textbf{0.223} & \textbf{0.457} & \textbf{0.091} & \textbf{0.528} & \textbf{0.102} & \textbf{0.631} \\
\midrule
\multirow{3}{*}{ResNet-50} 
    & Baseline     & 0.168 & 0.386 & 0.059 & 0.494 & 0.080 & 0.616 \\
ResNet-50 & CL       & 0.248 & 0.482 & 0.089 & 0.528 & 0.102 & 0.633 \\
    & HMCL   & \textbf{0.251} & \textbf{0.484} & \textbf{0.093} & \textbf{0.530} & \textbf{0.106} & \textbf{0.636} \\
\midrule
\rowcolor{gray!5}
\multicolumn{8}{l}{\textbf{ViT Models}} \\
\midrule
\multirow{3}{*}{ViT-Tiny} 
    & Baseline     & 0.150 & 0.365 & 0.054 & 0.487 & 0.077 & 0.612 \\
ViT Tiny & CL       & 0.304 & 0.536 & 0.099 & 0.541 & 0.102 & 0.637 \\
    & HMCL   & \textbf{0.310} & \textbf{0.542} & \textbf{0.105} & \textbf{0.546} & \textbf{0.110} & \textbf{0.642} \\
\midrule
\multirow{3}{*}{ViT-Small} 
    & Baseline     & 0.179 & 0.401 & 0.065 & 0.503 & 0.085 & 0.620 \\
ViT Small & CL       & \textbf{0.349} & \textbf{0.577} & 0.115 & 0.557 & 0.112 & 0.646 \\
    & HMCL   & 0.347 & 0.575 & \textbf{0.119} & \textbf{0.560} & \textbf{0.119} & \textbf{0.650} \\
\midrule
\multirow{3}{*}{ViT-Base} 
    & Baseline     & 0.171 & 0.392 & 0.064 & 0.499 & 0.084 & 0.619 \\
ViT Base & CL       & 0.\textbf{333} & 0.\textbf{564} & 0.115 & \textbf{0.556} & 0.113 & 0.646 \\
    & HMCL   & 0.324 & 0.555 & \textbf{0.116} & \textbf{0.556} & \textbf{0.118} & \textbf{0.648} \\
\midrule
\multirow{3}{*}{ViT-Large} 
    & Baseline     & 0.166 & 0.386 & 0.069 & 0.506 & 0.088 & 0.621 \\
ViT Large & CL       & \textbf{0.348} & \textbf{0.576} & 0.118 & 0.562 & 0.112 & 0.645 \\
    & HMCL   & 0.345 & 0.573 & \textbf{0.121} & \textbf{0.563} & \textbf{0.115} & \textbf{0.647} \\
\bottomrule
\end{tabular}
}
\end{table}

Table~\ref{tab:resnet_vit_results} presents the results obtained with the ResNet and ViT models. Baseline denotes the pretrained models without any additional finetuning on the patent domain. CL denotes the conventional contrastive learning approach defined in Equation~\ref{eq:ct} while HMCL indicates the proposed hierarchical multi-positive contrastive learning defined in Equation~\ref{eq:hct}. 

The proposed method is evaluated at all hierarchical levels (Patent ID, Subclass and Main Class), with each successive level containing incrementally more relevant images. The set of relevant items at the Patent ID level is a subset of those at the Subclass level, which also form a subset of the relevant items at the Main Class level. This explains why mAP may decrease as hierarchical levels increase (thus, results may be compared row by row, and not column by column). 

Overall, the hierarchical multi-positive contrastive loss enhances retrieval performance across all hierarchical levels. Notably, the proposed approach provides significant improvements with the ResNet architecture and lower parameter models such as ViT Tiny. While with larger ViT models, we notice improved performance at the Subclass and Main Class levels, we observe a slight deterioration at the Patent ID level. This trade-off is expected, as images from higher hierarchical levels have a higher similarity score and get higher in the ranking list. Also, we calculate the standard deviation between the runs, but we do not observe any significant difference between the methods. For the Patent ID level, the standard deviation is approximately $\pm 0.005$, for the Subclass level, it is $\pm 0.002$, and for the Main Class level, it is $\pm 0.001$ for both methods and metrics. 

Table~\ref{tab:clip_performance} reports the results with the CLIP model. First, we evaluate only the ViT component from a pretrained CLIP, in isolation from the language encoder. Additionally, we experiment in a multimodal setting with minimal language supervision where the textual descriptions are defined using the following format:
$$
\text{``This is a patent image of a [OBJECT\_NAME].''}
$$
where [OBJECT\_NAME] represents the object's description provided by DeepPatent2. These models provide significant improvements compared to the ViT and ResNet models from Table~\ref{tab:resnet_vit_results}. This can potentially being attributed to the extensive and contextualized pretrained phases of CLIP. Additionally, language supervision further improves performance. Finally, we observe a similar performance trade-off between Patent ID and the higher hierarchical levels, as previously shown in Table~\ref{tab:resnet_vit_results} for ViT Base and ViT Large. In the case of the CLIP models, the deterioration in performance at the Patent ID level is more pronounced, resulting from greater improvements in Subclass and Main Class levels.

\begin{table}[t]
\centering
\footnotesize
\caption{Performance of the CLIP models on patent image retrieval. All results are averaged across $3$ runs with different seeds. The top section reports the results when only the vision component from CLIP is used for training with no language supervision.}
\label{tab:clip_performance}
\scalebox{0.75}{
\begin{tabular}{>{\columncolor{gray!10}}l|l|cc|cc|cc}
\toprule
\rowcolor{gray!20}
\textbf{Model} & \textbf{Method} & \multicolumn{2}{c|}{\textbf{Patent ID}} & \multicolumn{2}{c|}{\textbf{Subclass}} & \multicolumn{2}{c}{\textbf{Main Class}} \\
& & \textbf{mAP} & \textbf{nDCG} & \textbf{mAP} & \textbf{nDCG} & \textbf{mAP} & \textbf{nDCG} \\
\midrule
\rowcolor{gray!5}
\multicolumn{8}{l}{\textbf{ViT component from CLIP}} \\
\midrule
\multirow{3}{*}{CLIP-B/16*} 
    & Baseline     & 0.179 & 0.400 & 0.077 & 0.565 & 0.077 & 0.563 \\
CLIP-B/16* & CL       & \textbf{0.373} & \textbf{0.596} & 0.120 & 0.609 & 0.121 & 0.609 \\
    & HMCL   & 0.356 & 0.582 & \textbf{0.128} & \textbf{0.613} & \textbf{0.139} & \textbf{0.618} \\
\midrule
\multirow{3}{*}{CLIP-L/14*} 
    & Baseline     & 0.213 & 0.437 & 0.088 & 0.577 & 0.087 & 0.574 \\
CLIP-L/14* & CL       & \textbf{0.454} & \textbf{0.663} & 0.136 & 0.626 & 0.135 & 0.624 \\
    & HMCL   & 0.452 & 0.661 & \textbf{0.146} & \textbf{0.633} & \textbf{0.155} & \textbf{0.634} \\
\midrule
\rowcolor{gray!5}
\multicolumn{8}{l}{\textbf{CLIP Models}} \\
\midrule
\multirow{3}{*}{CLIP-B/16} 
    & Baseline     & 0.179 & 0.400 & 0.077 & 0.565 & 0.077 & 0.563 \\
CLIP-B/16 & CL       & \textbf{0.401} & \textbf{0.619} & 0.128 & 0.617 & 0.125 & 0.613 \\
    & HMCL   & 0.386 & 0.608 & \textbf{0.137} & \textbf{0.623} & \textbf{0.148} & \textbf{0.625} \\
\midrule
\multirow{3}{*}{CLIP-L/14} 
    & Baseline     & 0.213 & 0.437 & 0.088 & 0.577 & 0.087 & 0.574 \\
CLIP-L/14 & CL       & \textbf{0.458} & \textbf{0.665} & 0.149 & 0.634 & 0.148 & 0.632 \\
    & HMCL   & 0.439 & 0.651 & \textbf{0.156} & \textbf{0.639} & \textbf{0.171} & \textbf{0.642} \\
\bottomrule
\end{tabular}
}
\end{table}

\begin{figure}[t]
    \centering
    \includegraphics[width=\linewidth]{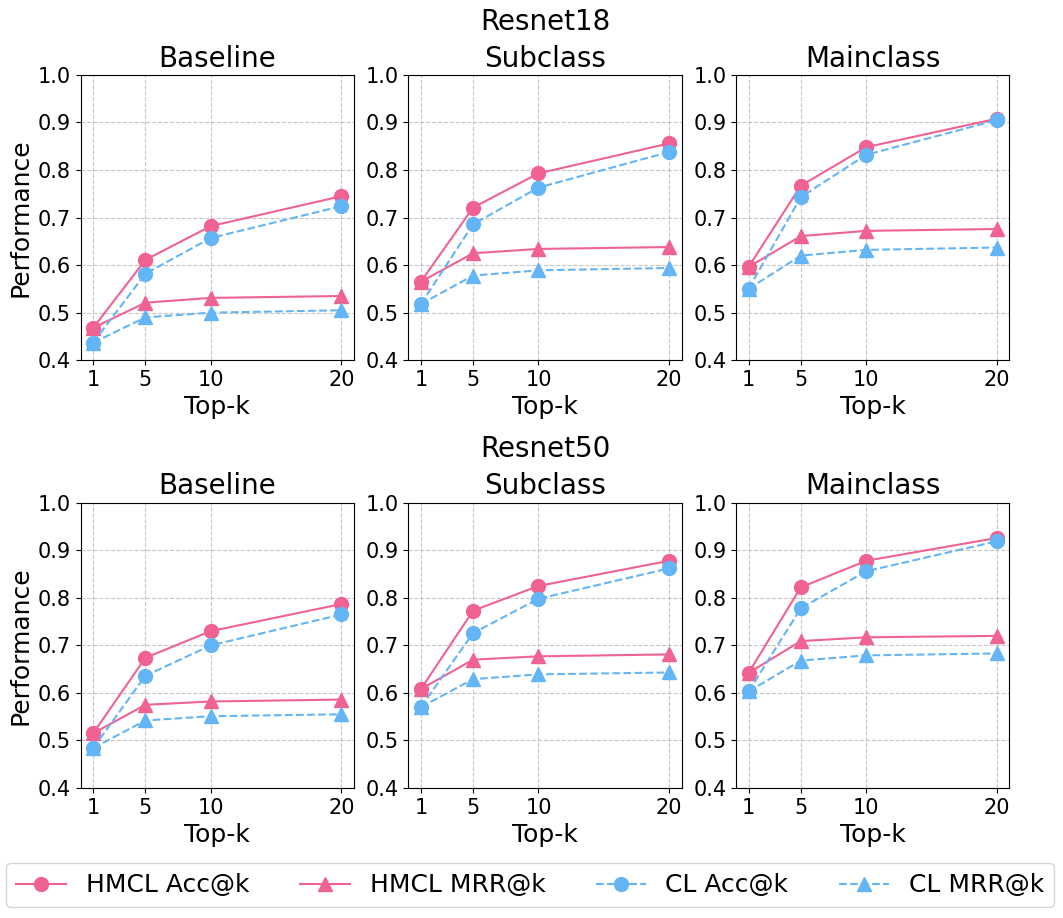}
    \caption{MRR@k and Acc@k for the Baseline, the conventional Contrastive Learning (CL) and the Hierarchical Multipositive Contrastive Learning (HMCL) method (for k = 1, 5, 10, and 20).}
    \label{fig:resnet_acc_mrr}
\end{figure}

Figure~\ref{fig:resnet_acc_mrr} reports the results with ResNet-18 and ResNet-50 using the metrics MRR@K and Acc@K for $K \in \{1, 5, 10, 20\}$, providing a more comprehensive overview of the retrieved list. For all levels (Patent ID, Subclass, and Main Class), the proposed approach outperforms the conventional contrastive learning method, with more relevant items being found at higher ranks in the retrieved list. 

\begin{figure}[th!] 
\centering 
\includegraphics[width=\linewidth]{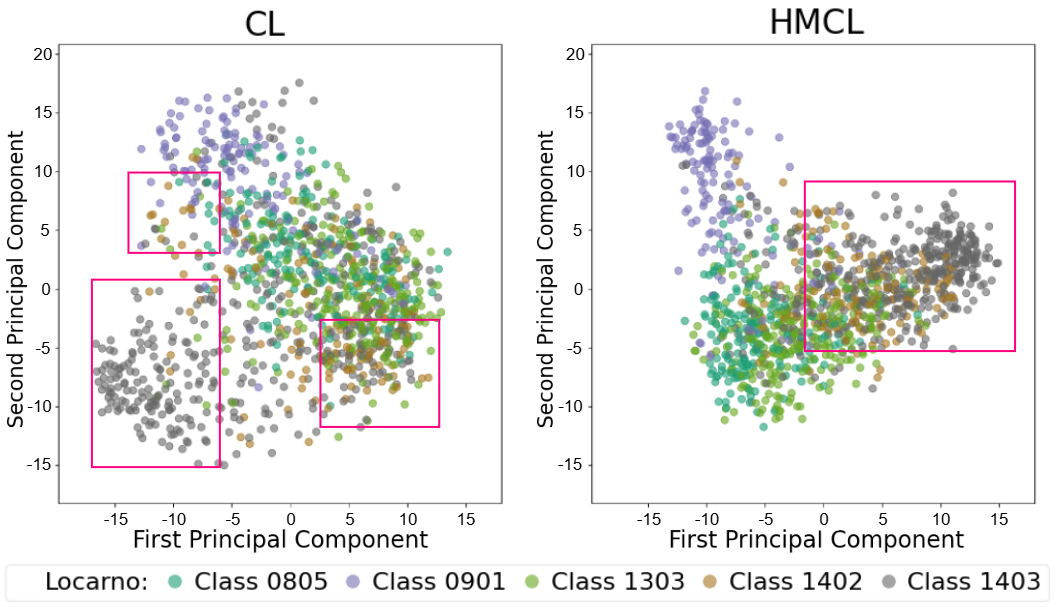} 
\caption{First and second principal component of the embeddings of 5 different subclasses from the LIC system for the conventional Contrastive Learning (CL) and the Hierarchical Multi-Positive Contrastive Learning (HMCL) method. The Subclass 1402 and 1403 from LIC belong to the same Main Class, while the rest to different ones. The red boxes enclose the main regions of Subclass 1402 and 1403.} 
\label{fig:pca_comparison} 
\end{figure}

Finally, we project the embeddings of ViT Base into $2$ dimensions using PCA. Figure~\ref{fig:pca_comparison} illustrates the samples from $5$ subclasses (where $2$ subclasses belong to the same main class). We notice that without any hierarchical information induced during training, the classes have a higher overlap and are less distinctly separated. In contrast, the proposed approach leads to more coherent clustering, with samples from the same subclass positioned closer together and subclasses of the same main class being closer in the embedding space.

\section{Conclusion}
\label{sec:conclusion}
In this paper, we presented a hierarchical multipositive contrastive learning approach to improve patent image retrieval. We integrated the hierarchical relationships of patents defined by the LIC system into the training process, allowing the models to capture this rich information in the embedding space. Our approach considers multiple positive pairs within a batch for an anchor image, with each pair being assigned a different relevance score, which reflects how closely their patents are classified within the chosen hierarchical taxonomy (e.g., LIC). Experimental results demonstrated that our approach enhanced performance at all hierarchical levels, exhibiting notable improvements with low parameter models. 

Our findings suggest that incorporating the hierarchical information of patents can improve patent image retrieval, opening several promising avenues for future research. One direction could be to explore hyperbolic embeddings, which are inherently more suitable for capturing hierarchical structures \cite{mettes2024hyperbolic}. Finally, our study was specifically focused on the LIC taxonomy. Future directions could investigate alternative taxonomies, for example the Cooperative Patent Classification system, which provides a more granular hierarchical structure with additional levels.

\bibliography{sample-ceur}

\clearpage
% \appendix

\end{document}